\documentclass[12pt]{article}

\usepackage{color}
\usepackage{algorithm}
\usepackage{algorithmic}
\usepackage{indentfirst}
\usepackage{subfigure}
\usepackage[left=2cm,right=2cm,top=2cm,bottom=3cm]{geometry}
\usepackage{booktabs}
\usepackage{graphicx}
\usepackage{tikz}
\usepackage{hyperref}
\usetikzlibrary{shapes,backgrounds,calc,arrows}
\usepackage{microtype}
\usetikzlibrary{positioning,shapes,shadows,arrows,calc,mindmap}

\title{Covariance Matrix Adaptation Evolution Strategy Assisted by Principal Component Analysis}
\author{Yangjie Mei, Hao Wang}
\date{LIACS, Leiden University, The Netherlands}
\providecommand{\keywords}[1]
{
  \small	
  \textbf{\textit{Keywords---}} #1
}

\begin{document}

\maketitle

\begin{abstract}
\noindent
Over the past decades, more and more methods gain a giant development due to the development of technology. Evolutionary Algorithms are widely used as a heuristic method. However, the budget of computation increases exponentially when the dimensions increase. In this paper, we will use the dimensionality reduction method Principal component analysis (PCA) to reduce the dimension during the iteration of Covariance Matrix Adaptation Evolution Strategy (CMA-ES), which is a good Evolutionary Algorithm that is presented as the numeric type and useful for different kinds of problems. We assess the performance of our new methods in terms of convergence rate on multi-modal problems from the Black-Box Optimization Benchmarking (BBOB) problem set and we also use the framework COmparing Continuous Optimizers (COCO) to see how the new method going and compare it to the other algorithms.

\end{abstract}
\keywords{Evolutionary Algorithm, Principal component analysis, Covariance Matrix Adaptation Evolution Strategy, COmparing Continuous Optimizers}

\section{Introduction \label{introduction}}
\hspace*{1cm}In the past several decades, more and more new methods are put forward with the development of social science and technology, more and more problems seem to have visible solutions. In this situation, evolutionary algorithms are widely used as a heuristic method to solve multi-objective optimization problems. However, though Evolutionary algorithms can find the optimum values or best solutions, the budget of computation increases exponentially with the increase of dimensions. \\
\hspace*{1cm}The curse of dimensionality lies in one of the difficulties of continuous optimization problems: when the dimension increases, it usually requires an exponentially increasing number of function evaluations to solve a problem for some target value. Hence, in this paper, we aim to delve into potential approaches to mitigate this issue, for which we plan to investigate the usefulness and effectiveness of various dimensionality reduction method PCA and couple it with an evolutionary algorithm CMA-ES. Supposedly, this integration of dimensionality reduction methods shall be implemented in an online manner, where the dimensionality reduction mapping and the inverse mapping thereof should be learned/adapted in each iteration of an evolutionary algorithm. What is more, compared to the traditional Evolutionary algorithm which fits well for the binary problem, we select the CMA-ES as the framework for our experiment. CMA-ES is a good particular kind of strategy for numerical optimization, it can adjust the iteration step automatically.\\
\hspace*{1cm}In section ~\textcolor{blue}{\ref{relatedwork}}, we will talk about some related work that already combines the PCA with Evolutionary Algorithm. In section~\textcolor{blue}{\ref{methods}}, we will talk about the core idea of dimensionality reduction methods and CMA-ES and how we combine the two methods. In section~\textcolor{blue}{\ref{Experiments}}, we gonna use the framework named CCOC to run our algorithms and run a benchmark for this. In section~\textcolor{blue}{\ref{discussion}}, results of the benchmark will be analyzed. In the last section~\textcolor{blue}{\ref{conclusions}}, we will conclude the results of the benchmark and figure out the advantages and disadvantages of our method and plan the future research.

\section{Related Work \label{relatedwork}}
\hspace*{1cm}In 2016, Dimitrios Kapsoulis~\cite{7850203} introduced the combination method of kernel PCA and Evolutionary algorithm. In a traditional evolutionary algorithm, we have three important steps during the iterations, Mutation, Crossover, and selection. Dimitrios Kapsoulis applied the K-PCA method before the mutation operation, lower the dimensions and invert mapping to the former dimension after the Crossover operation. With this method, the rate of convergence during the iterations can be rapid and it reduces the costs of computing while we have a huge multi-objective problem.\\
\hspace*{1cm}In 2020, Elena Raponi1, Hao Wang~\cite{raponi2020high} apply the PCA method on Bayesian Optimization (BO) and run the benchmark on COCO and assess the performance of PCA-BO in terms of the empirical convergence rate and CPU time on multi-modal problems from BBOB problem sets and gain some progress through the PCA method.

\section{Methods \label{methods}}
\subsection{Covariance matrix adaptation evolution strategy}
\hspace*{1cm}Before we talk about CMA-ES, we would like to introduce the traditional Evolutionary Algorithm, Genetic Algorithm.

\subsubsection{Genetic Algorithm}
\begin{figure}[!htbp]
\begin{center}
\includegraphics[height=10cm]{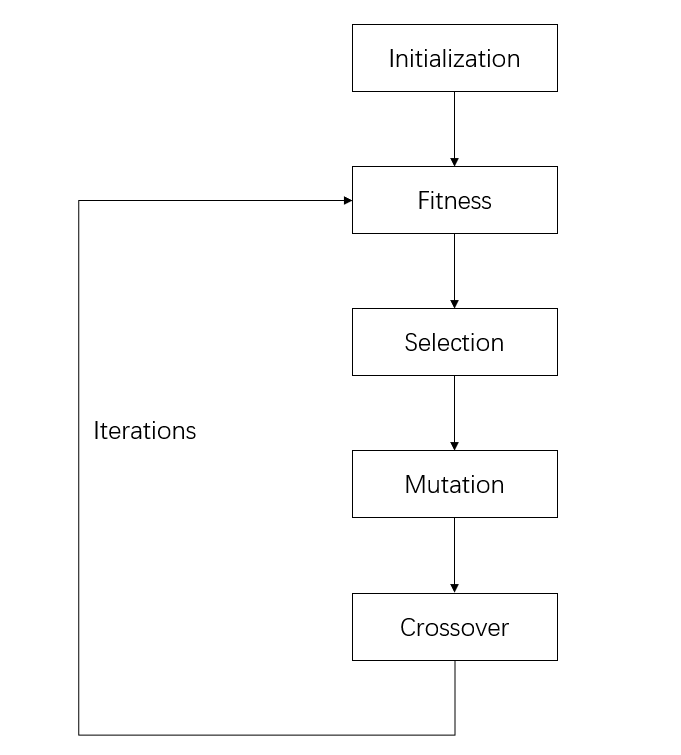}
\caption{The flow chart of Evolutionary Algorithm}
\label{fig:EA}
\end{center}
\end{figure}

\hspace*{1cm}Genetic Algorithm (GA) is a basic evolutionary Algorithm that is an optimized model that simulates Darwin's theory of biological evolution, it's a heuristic algorithm and was first proposed by Professor John Henry Holland in 1975. In genetic algorithm, each individual of the population is a feasible solution in the solution space, and the optimal solution is searched in the solution space by simulating the evolution process of organisms. And GA usually contains four parts core operations: Initialization, Mutation, Crossover, and Selection.The details of Evolutionary Algorithm are shown in Algorithm~\textcolor{blue}{\ref{alg:EA}} and Figure~\textcolor{blue}{\ref{fig:EA}}

\paragraph{Initialization}
The first populations are generated randomly with the given population size, and in each following generations, the populations are generated until it reaches its population size.
\paragraph{Mutation}
The mutation operation in genetic algorithm refers to the replacement of the gene values of some loci in the coding string of an individual chromosome with other alleles in the loci to form a new individual. Take this as an example, we got an existing individual like this: \textbf{00100}, and after mutation, we got a new individual like this: \textbf{00110}. The fourth loci of the individual have the mutation. Each locus in the individual has the mutation possibility $m_p$.
\paragraph{Crossover}
The crossover operation of genetic algorithm means that two paired chromosomes exchange some of their genes in a certain way, to form two new individuals.
For binary encoding individuals or floating-point encoding individuals, there are four kinds of Crossover methods
\begin{enumerate}
    \item \textbf{One-point Crossover}, it refers to setting only a random crossing point in the individual encoding string and then exchanging part of chromosomes of two paired individuals with each other.
    \item \textbf{Two-point Crossover and Multi-point Crossover}, Two or more random intersections were set in the individual coding string, and then two gene exchange to get two new individuals.
    \item \textbf{Uniform Crossover}, The genes in each locus of two paired individuals are exchanged with the same crossover probability, and forming two new individuals.
    \item \textbf{Arithmetic Crossover}, A linear combination of two individuals gives rise to two new individuals. This operation object is typically an individual encoded by a floating-point number.
\end{enumerate}

\paragraph{Selection}
In each problem, we have the fitness function. Like in some mathematics problems, we have the function and we want to find the lowest variables. Thus, each individual has it's own fitness. And then goes the Selection, the selection operation in genetic algorithm is used to determine how to select those individuals from the parent population in a certain way to pass on to the next generation population.
There are lots of Selection operations like Roulette Wheel Selection, Stochastic Tournament, and so on.

\begin{enumerate}
    \item \textbf{Roulette Wheel Selection} The probability of each individual entering the next generation is equal to the ratio of its fitness value to the sum of the fitness values of the individuals in the population. 
    \item \textbf{Stochastic Tournament} A pair of individuals are selected at each roulette wheel, and the two compete, with the fittest chosen, and so on until they are full.
\end{enumerate}

\begin{algorithm}
	\renewcommand{\algorithmicrequire}{\textbf{Input:}}
	\renewcommand{\algorithmicensure}{\textbf{Output:}}
	\caption{Evolutionary Algorithm}
	\label{alg:EA}
	\begin{algorithmic}[1]
		\REQUIRE Population size $P$, Crossover possibility $c_p$, Mutation possibility $m_p$, 
		\ENSURE Optimum result $r$
		
		\STATE Given population size $P$ and randomly generate the first offspring with a population $P$
		\STATE Generate the fitness of each individual and select the best top 10 percent of offspring as the elite
		\WHILE  {Not reach iteration limits or get optimum}
		    \STATE  Generate new 90 percent populations
    		\STATE  Randomly mutate the populations with possibility $m_p$
    		\STATE  Randomly crossover two population with possibility $c_p$
    		\STATE  Select the best 10 percent of offspring as the elite
		\ENDWHILE
		
		\STATE \textbf{return} Best Population 
	\end{algorithmic}  
\end{algorithm}

\subsubsection{Covariance Matrix Adaptation Evolution Strategy}
\hspace*{1cm}After talking about Genetic Algorithm, we will introduce the CMA-ES, which is quite good at generating numeric type solutions compared with GA. What is more, compared with the Simple Gauss evolutionary strategy (SGES), CMA-ES can adjust the step while training. Thus, the optimal value can be found in a short time with high accuracy.

In CMA-ES, the new distribution parameter can be like this:\\
\centerline{$\theta = (\mu,\sigma,C), p_\theta(x) \sim N(\mu,\sigma^2C) \sim \mu +{\sigma}N(0,C) $}\\
where $C$ is the covariance matrix and C has the following good properties:
\begin{enumerate}
    \item $C$ is always a diagonal matrix
    \item $C$ can be eigendecomposed into eigenvectors $B = [b_1,b_2,...,b_n]$ and eigenvalues $\lambda_1^2,\lambda_2^2,...,\lambda_n^2, D = diag(\lambda_1^2,...,\lambda_n^2)$
\end{enumerate}

\paragraph{Sample}
we can sample the new candidates in each generation by this formula:\\
\centerline{$x_i^{(t+1)}$ = $\mu^{(t)}$ $+$ $\sigma^{(t)}$$y_i^{(t+1)}$ where $y_i$$\sim$ $N(0,C^{(t)})$, $i = 1,...,\Lambda$}\\
where $x_i^{(t+1)}$ means the new candidates of the next generations, $\mu^{(t)}$ means the mean of elites of current generation and $\sigma^{(t)}$ means the step of current generation and $y_i$ is the subject to normal distributed.

\paragraph{Control Step}
We know that $\sigma^{(t)}$ controls the step of each generation, it is separated from the covariance matrix, so we can change the step size faster than we can change the full covariance. To estimate the step's appropriateness, CMA-ES get the sum of continuous moving sequence $\frac{1}{\lambda}\sum _i^\lambda y_i^{(j)},j=1,2,...,t$ to get the evolution path $p_\sigma$ and compare the evolution path with the path generated by random selection. if the evolution path is larger than the random selection path, reduce the $\sigma^{(t)}$, vice versa.\\

\paragraph{Adaptive covariance matrix}
The eigendecomposition of Covariance Matrix $C$ obeys:\\
\centerline{$C = BD^2B^T$},
We can re-estimate the origin Covariance Martix $C$ using the sampled population.\\ \centerline{$C_\lambda^{(t+1)} = \frac{1}{\lambda}\Sigma_{i=1}^{\lambda}y_i^{(t+1)}y_i^{(t+1)^T} = \frac{1}{\lambda\sigma^{(t)^2}}\Sigma_{i=1}^{\lambda}(x_i^{(t+1)}-\mu^{(t)})(x_i^{(t+1)}-\mu^{(t)})^T$}\\
In this formula, we can see that this estimation is only reliable when the population is larger. However, in each iteration, we want to has rapid iteration with a lower population. CMA-ES has a more reliable but more complicated way to update the $C$. It contains two kinds of unique ways.\\
The first method is using the history of $C$, we can also use the \textbf{Polyak Average}:\\
\centerline{$C^{(t+1)} = (1-\alpha_{c\lambda})C^{(t)} + \alpha_{c\lambda}C_\lambda^{(t+1)} =  (1-\alpha_{c\lambda})C^{(t)} + \alpha_{c\lambda}\frac{1}{\lambda}\Sigma_{i=1}^{\lambda}y_i^{(t+1)}y_i^{(t+1)^T}$},\\
and we choose the $\alpha_{c\lambda} = min (1,\lambda/n^2)$ normally.\\
The second way is using a path $p_c$ to log the symbol information, $p_c$ also subject to the normal distribution $N(0,C)$.\\
\centerline{$p_c^{(t+1)} = (1-\alpha_{cp})p_c^{(t)} + \sqrt{\alpha_{cp}(2-\alpha_{cp})\lambda}\frac{\mu^{(t+1)}-\mu^{(t)}}{\sigma^{(t)}}$},\\
and we use $p_c$ to update the covariance martix $C$:\\
\centerline{$C_\lambda^{(t+1)} = (1-\alpha_{c1})C^{(t)} + 1-\alpha_{c1}p_c^{(t+1)}p_c^{(t+1)^T}$}.\\
Finally, we combine these two methods and here we get the final update formula:\\
\centerline{$C^{(t+1)} = (1-\alpha_{c1}-\alpha_{c\lambda})C^{(t)} + \alpha_{c1}p_c^{(t+1)}p_c^{(t+1)^T} + \alpha_{c\lambda}\frac{1}{\lambda}\sum_{i=1}^{\lambda}y_i^{(t+1)}y_i^{(t+1)^T}$}

\subsection{Principal component analysis}
\hspace*{1cm}In main domain research, we need to monitor data with multi-variables and analyze it. Multi-variables can provide rich information but also bring the cost of computation and analysis. In continuous optimization problems, when the dimension increases, it usually requires an exponentially increasing number of function evaluations to solve a problem for some target value. Thus, reducing the variables but also keeping the information of variables seems to be an improvement of our research. Some methods can reduce the dimensions of data like Singular Value Decomposition (SVD), PCA, Latent Dirichlet Allocation (LDA), and so on. And we would choose PCA as our dimensionality reduction method.\\
\hspace*{1cm}The main idea of PCA is trying to map the N-dimension feature into K-dimension, K is smaller than N and this k dimension is a completely new orthogonal feature which is also called the principal component. PCA's job is to find a set of mutually orthogonal axes from the original space sequentially, and the choice of new axes is closely related to the data itself. Where, the selection of the first new coordinate axis is the direction with the greatest variance of the original data, the selection of the second new coordinate axis is the one with the greatest variance of the plane orthogonal to the first coordinate axis, and the third axis is the one with the greatest variance of the plane orthogonal to the first and second axes. And so on, PCA can get n of these axes. And PCA selects the first K axes as new axes. In fact, by calculating the covariance matrix of the data matrix, the eigenvalue eigenvectors of the covariance matrix are obtained, and the matrix is composed of the corresponding eigenvectors of k features with the largest eigenvalue (i.e., the largest variance) is selected. In this way, the data matrix can be transformed into a new space and the dimension reduction of data features can be realized.\\
\hspace*{1cm}Thus, the steps of PCA can be like this: Assuming that we have a dataset $X = \{x_1,x_2,...,x_n\}$ and we need to reduce its dimension into K, first, PCA will decentralize the data, which means each value has to minus the mean. Second, PCA will calculate the Covariance Matrix $C = \frac{1}{n}XX^T$ and get the eigenvalues and eigenvectors of $C$, sort the eigenvalues from large to small, and select the largest K of them. Then the corresponding K eigenvectors are taken as row vectors to form the eigenvector matrix $P$. Eventually, PCA gets the matrix $P$ and the mapping rule is $Y = PX$, $Y$ is the new dimension.

\subsection{CMA-ES with PCA}
\hspace*{1cm}In CMA-ES, all steps are done in the large dimensions, and what we want to do is try to reduce the dimensions dynamically during the iterations, with this method, we can reduce the cost of computation and also wipe the noise of some variables.\\
\hspace*{1cm} what we do in CMA-ES is that we apply PCA in the Sample operation, which is a sample process to get the population, of CMA-ES. And we don't do the PCA operation in the first iteration cause we need the first generation to get the PCA Matrix $P$. The pseudo-code description of CMA-ES has presented in Algorithm~\textcolor{blue}{\ref{alg:CMA-ES} }and CMA-ES-PCA core methodology works like this:\\ 
\hspace*{1cm} CMA-ES use Covariance Matrix $C^{old}$ to get the result of Sample operation. While CMA-ES-PCA get the new Covariance Matrix $C^{new}$ based on the Transform Matrix $P$ and $C^{old}$. Moreover, the Covariance Matrix computation subjects to two rules:\\ \centerline{\textbf{$Cov(x_1+x_2,y) = Cov(x_1,y)+Cov(x_2,y)$} and \textbf{$Cov(a*x_1,b*x_2) = a*b*Cov(x_1,x_2)$}}.\\
Here is an example that can help understand the process of the core idea: If the original dimension of the problem is 3 and now we have the Covariance Matrix $C^{old} {
\left[ \begin{array}{ccc}
c_{11}^{old} & c_{12}^{old} & c_{13}^{old}\\
c_{21}^{old} & c_{22}^{old} & c_{23}^{old}\\
c_{31}^{old} & c_{32}^{old} & c_{33}^{old}\\
\end{array} 
\right ]}$ and PCA Matrix $P {
\left[ \begin{array}{ccc}
1 & 0 & -1\\
-1 & 1 & 0\\
\end{array} 
\right ]}$ which can transform the variables from 3 variables($x_1, x_2, x_3$) into 2 variables($y_1, y_2$):
\centerline{${
\left[ \begin{array}{c}
y_1 \\
y_2\\
\end{array} 
\right ]}$ = $P {
\left[ \begin{array}{ccc}
1 & 0 & -1\\
-1 & 1 & 0\\
\end{array} 
\right ]}$ $\times$ ${
\left[ \begin{array}{c}
x_1 \\
x_2\\
x_3\\
\end{array} 
\right ]}$.} And with the given PCA Matrix $P$, we can get the following equation:
{\color{red}{$y_1 = x_1 - x_3$}} and {\color{red}{$y_2 = -x_1 + x_2$}}. Therefore, we can get the new corvariance between $y_1$ and $y_2$:$Cov(y_1,y_2)$:
\begin{enumerate}
    \item {\color{red}{$Cov(y_1,y_2) = Cov(x_1- x_3 , -x_1 + x_2)$}}
    \item {\color{red}{$Cov(x_1 - x_3 , -x_1 + x_2) = -Cov(x_1,x_1) + Cov(x_1,x_3) + Cov(x_1,x_2) - Cov(x_1,x_2)$}}
    \item {\color{red}{$Cov(y_1,y_2) = -c_{11}^{old} + c_{13}^{old} + c_{12}^{old} - c_{12}^{old} = c_{13}^{old} -c_{11}^{old}$  }}
    \item {\color{red}{$Cov(y_1,y_2) = c_{13}^{old} -c_{11}^{old}$  }}
\end{enumerate}
And for normal situation, if we want to get the new Convariance Matrix $C^{new}$, 
with the given PCA Matrix $P_{m{\times}n}$ and old Convaraince Matrix $C^{old}$. The rule of getting new Corvariance Matirx $C_{i,j}^{new}$ can be like this: \\
\centerline{{\color{blue}{$C_{i,j}^{new} = \sum_{a=1}^n\sum_{b=1}^nP_{i,a}P_{j,b}Cov_{a,b}^{old}, i,j<=m$}}}
After we get the new Corvariance Matrix $C^{new}$, we can sample the new solutions by\\
\centerline{{\color{blue}{$x_i^{(t+1)}$ = $\mu^{(t)}$ $+$ $\sigma^{(t)}$$y_i^{(t+1)}$ where $y_i$$\sim$ $N(0,C_{new}^{(t)})$}}} and we get some solutions $Y_{new}$, we get the pseudo-inverse matrix $P^{T}$ and the former $X = P^{T}Y_{new}$

\hspace{1cm} Here we add the PCA method into the CMA-ES sample operation to reduce the dimension and also reduce the cost of computation and also wipe some potential noise. The complete algorithm can be like this~\textcolor{blue}{\ref{alg:CMA-ES}} and we can see how it going in the experiment.
\begin{algorithm}
	\renewcommand{\algorithmicrequire}{\textbf{Input:}}
	\renewcommand{\algorithmicensure}{\textbf{Output:}}
    	\caption{Covariance Matrix Adaptation Evolution Strategy with Dimensionality Reduction Method}
	\label{alg:CMA-ES}
	\begin{algorithmic}[1]
		\REQUIRE    Learning rates: $\alpha_\mu$, $\alpha_\sigma$, $\alpha_{cp}$, $\alpha_{c1}$, $\alpha_{c\lambda}$
		\REQUIRE    Generation Count: t = 0
		\REQUIRE    Attenuation Factor: {$d_\sigma$}
		\REQUIRE    Evolutionary Paths: $p_\sigma^{(0)}$ = 0, $p_c^{(0)}$ = 0
		\REQUIRE    Default Covariance Matrix: $C^{(0)}$ = $I$
		
		\ENSURE     $\mu^{(t)},\sigma^{(t)},C^{(t)}$
		\STATE  Sample $x_i^{(1)}$ = $\mu^{(0)}$ $+$ $\sigma^{(0)}$$y_i^{(1)}$ where $y_i$$\sim$ $N(0,C^{(0)})$, $i = 1,...,\Lambda$ 
		\STATE  Select top $\lambda$ samples with the best fitness $x_i^{(t+1)}$, $i = 1,...,\lambda$ 
		\STATE  Update the parameters
		\WHILE  {Not hit stopping criteria}
		    \STATE  Apply the PCA methods to the existing elite solutions and get the PCA Matrix $P$
		    \STATE  Transform the Covariance Matrix from $C_\Lambda^{(t)}$ to $C_\Theta^{(t)}$ with the existing Matrix $P$ and $C_\Lambda^{(t)}$
		    \STATE  Sample $x_i^{(t+1)}$ = $\mu^{(t)}$ $+$ $\sigma^{(t)}$$y_i^{(t+1)}$ where $y_i$$\sim$ $N(0,C^{(t)})$, $i = 1,...,\Theta$
		    \STATE  invert-map the Samples from lower Dimension back to former Dimension 
    		\STATE  Select top $\lambda$ samples with the best fitness $x_i^{(t+1)}$, $i = 1,...,\lambda$ 
    		\STATE  $\mu^{(t+1)} \leftarrow \mu^{(t)} + \alpha_\mu\frac{1}{\lambda}\sum_{i=1}^\lambda(x_i^{(t+1)}-\mu^{(t)}) $
    		\STATE  $p_\sigma^{(t+1)} \leftarrow (1-\alpha_\sigma)p_\sigma^{(t)} + \sqrt{\alpha_\sigma(2-\alpha_\sigma)\lambda}C^{(t)^{-\frac{1}{2}}}\frac{\mu^{(t+1)}-\mu^{(t)}}{\sigma{(t)}}$
    		\STATE  $\sigma^{(t+1)} \leftarrow \sigma^{(t)}exp(\frac{\alpha_\sigma}{d_\sigma}(\frac{||p_\sigma^{(t+1)}||}{E||N(0,I)||}-1))$
    		\STATE  $p_c^{(t+1)} \leftarrow (1-\alpha_{cp})p_c^{(t)} + \sqrt{\alpha_{cp}(2-\alpha_{cp})\lambda}\frac{\mu^{(t+1)}-\mu^{(t)}}{\sigma{(t)}}$
    		\STATE  $C^{(t+1)} \leftarrow (1-\alpha_{c1})C^{(t)} + \alpha_{c1}p_c^{(t+1)}p_c^{(t+1)^T}   $
    		\STATE  $C^{(t+1)} \leftarrow (1-\alpha_{c1}-\alpha_{c\lambda})C^{(t)} + \alpha_{c1}p_c^{(t+1)}p_c^{(t+1)^T} + \alpha_{c\lambda}\frac{1}{\lambda}\sum_{i=1}^{\lambda}y_i^{(t+1)}y_i^{(t+1)^T}$
    		\STATE $t+1 \leftarrow t$
    		
		\ENDWHILE
		
		\STATE \textbf{return} $\mu^{(t)},\sigma^{(t)},C^{(t)}$
	\end{algorithmic}  
\end{algorithm}

\section{Experiments \label{Experiments}}
\textbf{Experiment Setup.}  We assess the CMA-ES-PCA method on ten multi-modal functions which are from the BBOB problem set~\cite{brockhoff2016using} and compare its performance with CMA-ES and CMA-ES-PCA-Randomly (use PCA method randomly in iteration). The three algorithms are both tested on Functions F15-F24. F15-F19 are multi-modal functions with adequate global structure while F20-F24 are multi-modal functions with weak global structure. We would see how three algorithms work in these functions. What is more, we test three algorithms on three different dimensions: 10, 20, 30, and with several budgets equal \textbf{20D} function evaluations. For accuracy, we repeat each algorithm on each function 30 times. The experiment is worked on an Intel Intel® Core™ i7-9700K Processor @ 3.6GHz.\\

\begin{figure}[htbp]
\centering

\includegraphics[width=0.7\textwidth]{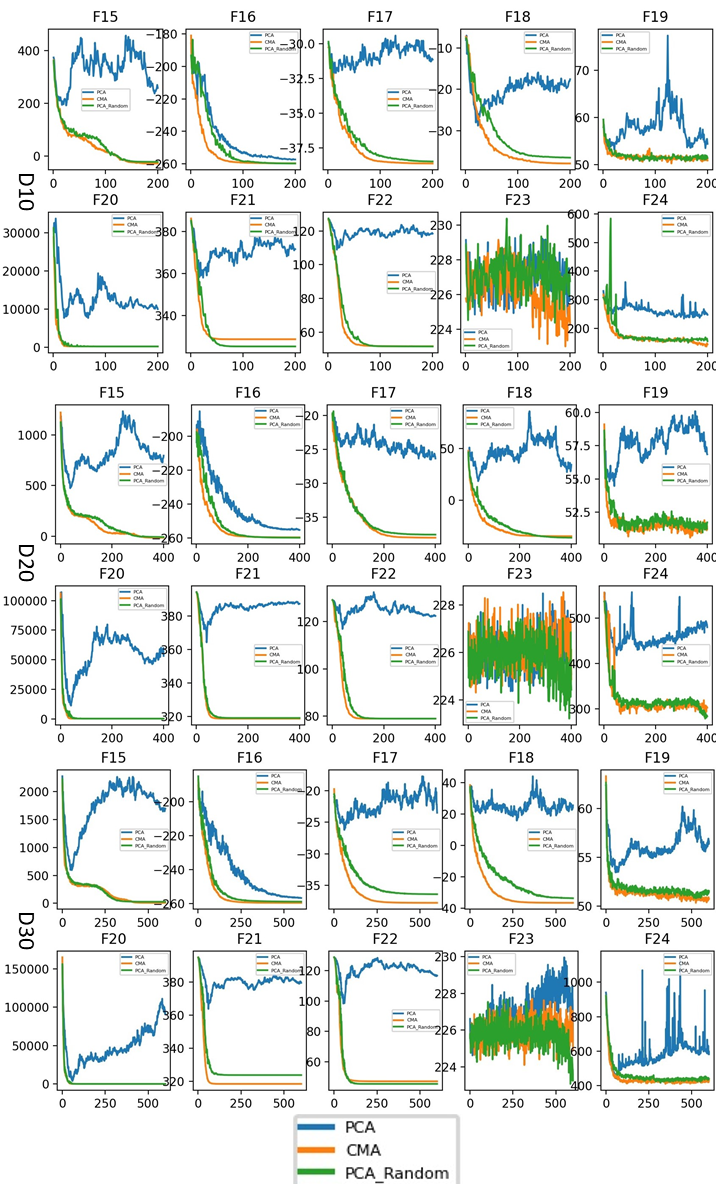}

\caption{This figure shows the convergence rate of three methods: CMA-ES, CMA-ES-PCA, and CMA-ES-PCA-Randomly. The X-axis indicates the budget while the Y-axis shows the evaluation function value, lower is better. The benchmark is done on Dimension 10, 20, and 30 and does the test on 10 multi-modal functions F15-F24.}
\label{fig.Result}
\end{figure}

\begin{figure}[htbp]
\centering
\subfigure[DSA]{
\includegraphics[width=7cm]{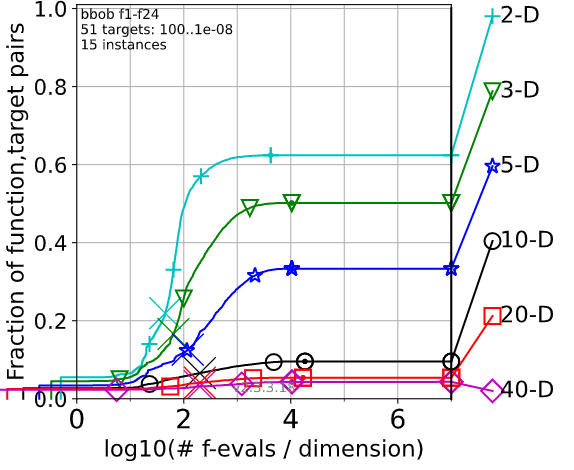}

}
\quad
\subfigure[CMA-ES]{
\includegraphics[width=7cm]{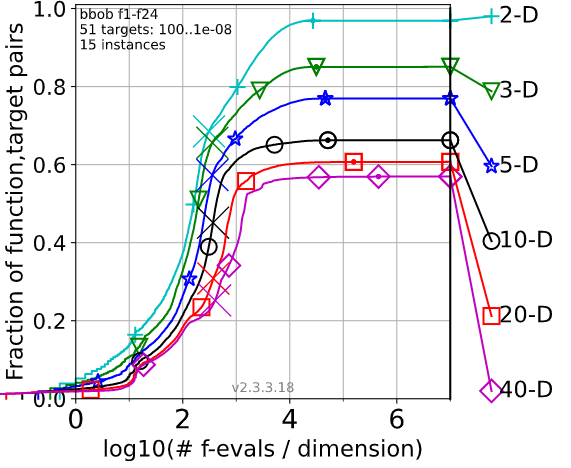}
}
\quad
\subfigure[CMA-ES-PCA]{
\includegraphics[width=7cm]{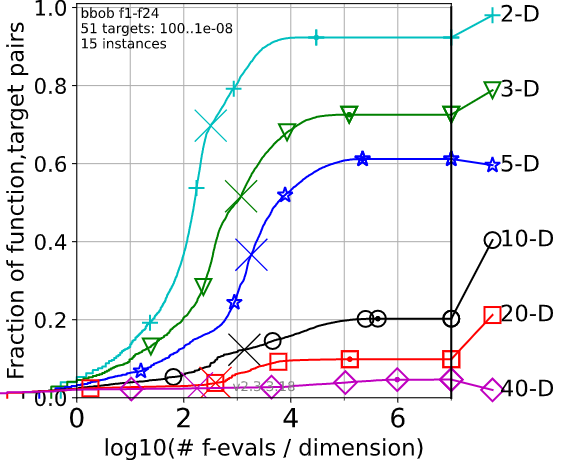}
}
\quad
\subfigure[CMA-ES-PCA-Randomly]{
\includegraphics[width=7cm]{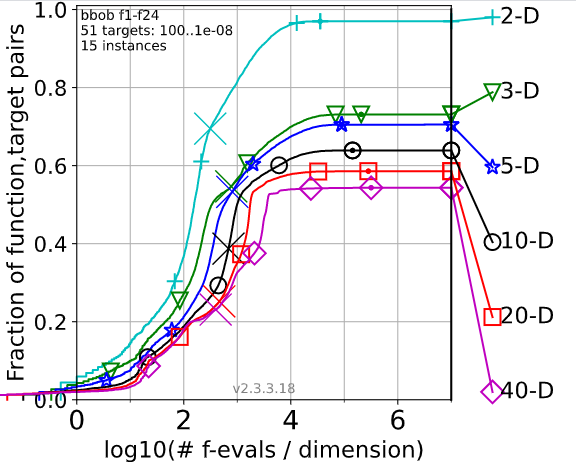}
}
\caption{These four figures show the summary of four algorithms' Empirical cumulative distributions (ECDF) of run lengths and speedup ratios in 2D (top) to 40D (bottom) overall functions. The upper left figure is the Downhill Simplex Algorithm, the upper right is the CMA-ES Algorithm and the lower left is the CMA-ES algorithm with PCA method and the lower right is the CMA-ES algorithm with a possibility (0.5) to use PCA The angle of inclination shows the rate of convergence and the Y-axis shows the performance in the target, higher means better.}
\label{overall}
\end{figure}

\begin{figure}[htbp]
\centering
\subfigure[DSA]{
\includegraphics[width=8cm]{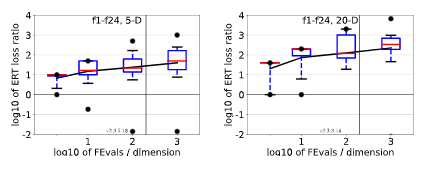}

}
\quad
\subfigure[CMA-ES]{
\includegraphics[width=8cm]{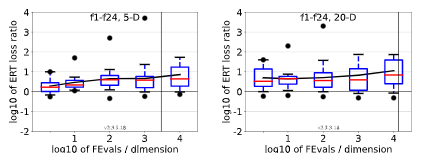}
}
\quad
\subfigure[CMA-ES-PCA]{
\includegraphics[width=8cm]{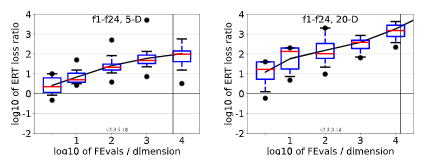}
}
\quad
\subfigure[CMA-ES-PCA-Randomly]{
\includegraphics[width=8cm]{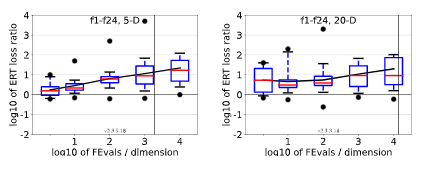}
}
\caption{These four figures show the ERT loss ratio versus the budget in the number of f-evaluations divided by dimension. The upper left figure is the Downhill Simplex Algorithm, the upper right is the CMA-ES Algorithm and the lower left is the CMA-ES algorithm with PCA method and the lower right is the CMA-ES algorithm with a possibility (0.5) to use PCA. The higher the means the algorithm performs worse in the benchmarking.}
\label{ERT}
\end{figure}

\hspace{1cm}What is more, we select four different methods and run in on the COmparing Continuous Optimizers (COCO)~\cite{nikolaus_hansen_2019_2594848}, we select 2009 BBOB problem set and test four algorithms, downhill simplex algorithm, CMA-ES, CMA-ES-PCA and CMA-ES-PCA-Randomly on all functions F1-F24 and the dimensions we set are 2, 3, 5, 10, 20  and 40D. Here is the results of three methods: Figure.\textcolor{blue}{\ref{fig.Result}}, Figure.\textcolor{blue}{\ref{overall}}, Figure.\textcolor{blue}{\ref{ERT}} and Table.\textcolor{blue}{\ref{ERT_LOSS_RATIOs}}

\begin{table}[!htbp]
  \begin{center}
    \begin{tabular}{ccccccccc}
      \toprule
        Algorithm&$\#$FEs/D& Dimension&best& 10$\%$ & 25$\%$ & median & 75$\%$ & 90$\%$  \\
      \midrule
      CMA-ES &2 &5& 0.55&	0.65&	0.94&	1.7	&3.0&	10    \\
      &10 &5& 1.1&	1.2	&1.7&	2.2	&3.9&	10  \\
      &100&5& 0.46&	1.2&	2.0&	4.0&	6.6&	18   \\
      &1e3 &5& 0.58	&0.92&	1.6	&3.8&	6.0	&30    \\
      &1e4 &5& 0.73&	0.94&	1.8	&4.4&	19&	58    \\
      &$RL_{US}/D$&5&	80&	2e2	&3e2&	4e2	&6e2&	8e2\\
      \bottomrule
      CMA-ES-PCA&2 &5& 0.47   & 0.81   & 1.1   & 2.2&  7.5 & 10    \\
      &10 &5& 2.6&	3.1	&3.7&	5.0	&11	&20  \\
      &100&5& 3.8 &	10&	15&	21	&32	&1.4e2   \\
      &1e3 &5& 7.2	&20	&32	&47	&89	&1.7e2     \\
      &1e4 &5& 3.2	&8.9&	40&	97&	1.5e2&	8.1e2     \\
      &$RL_{US}/D$&5&	70&	8e2&	9e2&	2e3&	4e3	&6e3\\
      \bottomrule
      CMA-ES-PCA-Randomly&2 &5&0.61&	0.64&	0.91&	1.6&	2.5&	10    \\
      &10 &5& 0.69&	1.1&	1.7&	2.2&	3.8&	10  \\
      &100&5& 	0.62&	2.0&	3.8&	6.1	&8.2&	18  \\
      &1e3 &5& 0.65	&1.4&	3.2	&8.9&	32&	1.1e2 \\
      &1e4 &5&1.00&	2.3&	4.5&	17	&55	&2.0e2    \\
      &$RL_{US}/D$&5&	60&	5e2	&8e2&	8e2&	1e3&	1e3\\
      \bottomrule
      CMA-ES&2 &10& 0.59&	0.97&	1.8&	3.3&	14&	40   \\
      &10 &10& 0.64&	1.1	&2.3	&4.5&	6.2	&27  \\
      &100&10& 0.56	&0.97&	1.5	&3.6&	10&	49   \\
      &1e3 &10& 0.48&	0.76&	1.2	&3.9&	20	&89    \\
      &1e4 &10& 0.48&	0.76&	2.3&	6.9&	43&	79    \\
      &$RL_{US}/D$&10&1e2&	2e2&	3e2&	4e2&	7e2&	1e3\\
      \bottomrule
      CMA-ES-PCA&2 &10& 0.59&	1.1&	4.9	&17	&40	&40   \\
      &10 &10& 4.8&	7.1&	15&	1.3e2&	2.0e2&	2.0e2  \\
      &100&10& 9.6&	17&	59&	1.0e2&	3.3e2&	2.0e3   \\
      &1e3 &10& 64&	66&	1.8e2&	3.8e2&	5.1e2&	2.8e3     \\
      &1e4 &10& 2.2e2&	3.7e2&	7.2e2&	1.5e3&	2.8e3&	2.4e4     \\
      &1e5 &10& 1.0e3&	2.2e3&	3.7e3&	9.7e3&	2.8e4&	2.4e5     \\
      &$RL_{US}/D$&10&30&	1e2	&2e2&	4e2	&1e3&	1e4\\
      \bottomrule
      CMA-ES-PCA-Randomly&2 &10& 0.68&	0.81&	1.3&	5.3	&27	&40   \\
      &10 &10& 0.56&	1.2&	2.1	&3.1&	6.0&	2.0e2  \\
      &100&10& 0.24&	1.1	&2.9&	3.9	&8.5&	48   \\
      &1e3 &10& 0.74&	1.0&	2.4&	9.2	&30	&72     \\
      &1e4 &10& 0.59&	1.5&	2.9&	9.1&	75	&1.3e2     \\
      &$RL_{US}/D$&10&60&	3e2	&3e2&	5e2&	6e2&	2e3\\
      \bottomrule

    \end{tabular}
    \caption{The table shows the ERT loss ratios of three CMA-ES algorithms in all datasets and problems with dimensions 5D and 20D. The last row $RL_{US}/D$ gives the number of function evaluations in unsuccessful runs divided by dimension. Shown are the smallest, 10$\%$-ile, 25$\%$-ile, 50$\%$-ile, 75$\%$-ile, and 90$\%$-ile value (smaller values are better). \label{ERT_LOSS_RATIOs}}
    
  \end{center}
\end{table}

\section{Discussion \label{discussion}}
\hspace{1cm} From the result, we can some interesting characteristics of the CMA-ES and also our new methods. From the Figure~\textcolor{blue}{\ref{fig.Result}} we can see that in all dimensions and most functions, the CMA-ES-PCA presents a precarious situation, sometimes it cannot convergence at the end of iterations like F15, F17, F18, and so on. However, CMA-ES can also have a great performance in Function 16. What is more, we can also see some good parts of CMA-ES-PCA, in Function 18 with dimension 10, CMA-ES-PCA has a faster convergence rate at the beginning of iterations, which indicates that the CMA-ES-PCA has the potential to get a better result and faster convergence rapid during the iterations, which can maybe also useful in other traditional problems. Therefore, we add another method means CMA-ES-PCA-Randomly, which means we don't apply the PCA method at each iteration, we operate the operation randomly (0.5 in this experiment), and the results are pleasantly surprising. The CMA-ES-PCA-Randomly can have equal performance compared to CMA-ES, what is more, it has a better performance in  Function 21, the algorithm gains a better result in the final iterations. And this algorithm performs better in multi-modal functions with a weak global structure compared to an adequate global structure. And in Function 23, CMA-ES performs worse and also CMA-ES-PCA cannot perform well either.\\

\hspace{1cm}From the Figure~\textcolor{blue}{\ref{overall}} we can see that the CMA-ES performs well overall targets, but CMA-ES-PCA performs worse compared to CMA-ES, but still better than Downhill Simplex Algorithm (DSA), and we can see the CMA-ES-PCA performs equally well at 2-Dimension, what causes this is that we don't do the PCA method at 2-D, cause when transforming the variables from two to one, we cannot get the covariance matrix, which at least needs two variables. But CMA-ES-PCA performs not bad at 3-D and 5-D, what causes this could be that the PCA-method doesn't change much, it doesn't like 10-D, 20-D, and 40-D, which may change the dimensions from 40-D to 10-D, the changes are too big and when transfer back the former dimensions, the loss can be huge. And what causes the algorithm to perform badly in 10-D, 20-D and 40-D is that at the beginning of Iterations, the PCA method gets the new Covariance Matrix which has a big deviation from the old Covariance Matrix, and it's hard for it to amend the following iterations. Thus, we make a new method that uses the PCA method randomly in the iterations, we set the possibility equals 0.5, which means the PCA method has the 50 percent possibility to operate in the Sample method. From the result, we can see that the method doesn't change much in 3-D and 5-D, but the method gets a giant improvement in the 10-D, 20-D, and 40-D problems, which means this method can do well in high dimensions when compared to lower dimensions. What is more, the inclination of a line is more rapid compared to the CMA-ES, which indicates the CMA-ES-PCA can gain a rapid rate of convergence. \\

\hspace{1cm}From the Table~\textcolor{blue}{\ref{ERT_LOSS_RATIOs}}, we can see that in the early phase of convergence like ($\#$FEs/D = 10), the CMA-ES with PCA performs not bad compared to CMA-ES, but it performs worse in the latter phase of convergence, especially in high dimension. But we can also see that CMA-ES-PCA and CMA-ES-PCA (randomly)s' best $RL_{US}/D$ are better than CMA-ES, which means when we apply the PCA method, we can lower unsuccessful runs in some specific time, the PCA methods brings instability but also performs better in some time. What is more, if we just use PCA randomly, we can gain both good performance and stability. This can be also seen in Figure \textcolor{blue}{\ref{ERT}}, the CMA-ES-PCA has a larger span compared to other algorithms. And CMA-ES-PCA (randomly) gains a better performance in higher dimensions.

\section{Conclusions and Further Research \label{conclusions}}
\hspace{1cm}We can see that CMA-ES-PCA-Randomly can gain both good performance and stability when compared to the CMA-ES, and it performs well in high dimensions, and it can also gain a better result like Function 21, Function 22 in BBOB cause it's randomness, which means we can use it in some practical problem. And why CMA-ES-PCA performs worse in the benchmark is reasonable, the PCA method will reduce the dimension, which wipes the noise, reduces time cost, but we also need to transfer back to the former dimension, which means we must lose some characteristics of the variables, and this cause the instability and better result, but it also brings worse result in the main time. However, if we use the possibility to use the PCA randomly, or we can use some more intelligent way to use the PCA, the results perform much better than we thought. Thus, it's doable to apply the PCA-method randomly in the iterations of CMA-ES. \\

In further research, we would like to do in four points:

\begin{enumerate}
    \item Add more other Dimensionality reduction methods like Random Forest, T-SNE, and Autoencoders.
    \item Change the parameters of PCA to get better performance.
    \item Change the possibility to apply the PCA method to see its change.
    \item Use a more intelligent algorithm to select the PCA-method in the iterations of CMA-ES.
\end{enumerate}

\bibliographystyle{plain}
\bibliography{bibliography}


\end{document}